\documentclass[manuscript]{acmart}
\usepackage{geometry}
\usepackage{amsmath,amsfonts}
\usepackage{enumitem}
\usepackage{algpseudocode}
\usepackage{textcomp}
\usepackage{xcolor}
\usepackage{color}
\usepackage{graphicx}
\usepackage{pifont}
\newcommand{\xmark}{\ding{55}}%
\DeclareUnicodeCharacter{2265}{$\geq$}
\usepackage[ruled,linesnumbered,noresetcount]{algorithm2e}
\begin{document}

\title{Learning with Noisy Labels for Human Fall Events Classification:\\ Joint Cooperative Training with Trinity Networks}
\author{Leiyu Xie}
\authornotemark[1]
\email{l.xie6@newcastle.ac.uk}
\orcid{0000-0001-7023-2593}
\author{Yang Sun}
\authornotemark[2]
\email{yang.sun@bdi.ox.ac.uk}
\author{Syed Mohsen Naqvi}
\authornotemark[1]
\email{mohsen.naqvi@newcastle.ac.uk}

\affiliation{%
  \institution{Newcastle University}
  \city{Newcastle Upon Tyne}
  \authornotemark[1]
  \country{UK}
  \postcode{NE1 7RU}
}
\affiliation{%
  \institution{University of Oxford}
  \city{Oxford}
  \authornotemark[2]
  \country{UK}
  \postcode{OX3 7LF}
}


\begin{abstract}
With the increasing ageing population, fall events classification has drawn much research attention. In the development of deep learning, the quality of data labels is crucial. Most of the datasets are labelled automatically or semi-automatically, and the samples may be mislabeled, which constrains the performance of Deep Neural Networks (DNNs). Recent research on noisy label learning confirms that neural networks first focus on the clean and simple instances and then follow the noisy and hard instances in the training stage. To address the learning with noisy label problem and protect the human subjects' privacy, we propose a simple but effective approach named Joint Cooperative training with Trinity Networks (JoCoT). To mitigate the privacy issue, human skeleton data are used. The robustness and performance of the noisy label learning framework is improved by using the two teacher modules and one student module in the proposed JoCoT. To mitigate the incorrect selections, the predictions from the teacher modules are applied with the consensus-based method to guide the student module training. The performance evaluation on the widely used UP-Fall dataset and comparison with the state-of-the-art, confirms the effectiveness of the proposed JoCoT in high noise rates. Precisely, JoCoT outperforms the state-of-the-art by 5.17\% and 3.35\% with the averaged pairflip and symmetric noises, respectively.
\end{abstract}

%

\ccsdesc[500]{Computing Methodologies~Artificial Intelligence}
\ccsdesc[300]{Computing methodologies~Machine learning algorithms}
\ccsdesc[200]{Computing methodologies~Modeling and simulation}
\ccsdesc[100]{Computing methodologies~Computer vision}

\keywords{Fall Detection, Artificial Intelligence, Deep Learning, Multi-class Fall Events, Healthcare, Noisy Label Learning, Assisted Living, Smart Home, Data Corruption}


\maketitle

\section{Introduction}
With the increasing ageing population across the globe, falls have surpassed cardiovascular diseases and cancer, becoming the major issue leading to elderly people's death \cite{wang2020elderly, 6279483}. Moreover, bodily harm, such as fractures, and even mental harm from falls will lead to secondary injuries in elderly people \cite{xu2018new}. According to a report of the World Health Organization, 28\% to 35\% of people aged 65 or over fall each year, and the number increases to 32\% to 42\% when people are at the age of 70 or over \cite{world2008global}. As the age of the people increases, these numbers get even worse. As reported in \cite{burns2018deaths}, the increasing rate of the ageing population's death caused by falls is approximate 3.0\% each year between 2007 to 2016 in the USA \cite{safari2018cross}. Meanwhile, the ageing population has been increasing significantly worldwide in recent years. Therefore, assisted living systems are required. There are many types of fall events in the real-world living environment, such as falling to a chair, falling when going upstairs, falling using hands, falling with knees and other fall events \cite{mubashir2013survey, xie2022privacy, xie2021skeleton}. Most of them share common characteristics but still have distinctive differences. Different types of fall events will lead to various injuries in human bodies, which have different harm levels \cite{xie2021skeleton}. For example, elderly people will suffer head injuries from falling down stairs events and back injuries from falling backwards events. Conventional fall detection only focuses on whether the fall event has occurred but ignores different types of fall events, which may cause different injuries to elderly people \cite{ren2019research, ramachandran2020survey}. A detailed classification of the types of fall events can provide rescuers with timely information on the decision for the type of resuscitation equipment to best avoid secondary injuries that elderly people may suffer. Furthermore, provide them with targeted and efficient medical care. Therefore, a robust fall classification system is needed for better healthcare. 

According to the approaches applied in recent years, we divide fall detection methods mainly into two categories: wearable sensors and non-wearable sensors based approaches. In\cite{bagala2012evaluation}, 13 different algorithms based on accelerometers are included. The algorithm with the highest performance achieves 98\% for fall detection. In \cite{tamura2009wearable}, researchers proposed a system which contains an airbag for use when a fall is detected. However, there are two issues with the wearable-based approaches. Firstly, elderly people may be reluctant or forget to wear the equipment for long periods \cite{delahoz2014survey}. Secondly, some of these sensors are battery-powered, and elderly people may neglect the power issue. The most popular method used in non-wearable sensor based approaches is the vision-based method, one of the most common sensors is the surveillance camera \cite{ramirez2021fall, gutierrez2021comprehensive, harrou2017vision, espinosa2019vision}. It is considered as one of the sub-classes of context-aware systems and computer vision techniques are widely used to meet the requirements of non-wearable sensors as opposed to wearable sensors-based approaches. A shape-matching algorithm is proposed in \cite{rougier2011robust} to detect human fall events in the video clip. A deep learning-based approach which uses multiple cameras and convolutional neural networks is proposed for better fall detection in \cite{espinosa2019vision}. This type of approach could continuously and automatically be used for fall classification without the limitations of wearable sensor-based approaches. Although such vision-based methods have the advantages mentioned above, one of the most concerning issues is privacy protection \cite{6566012}. The privacy mitigation in the proposed work includes the subject's personal identity and background (contextual) information protection, which can be easily compromised in RGB images. Therefore, human skeleton data which only contains the key body landmarks and completely removes background (contextual) information is used in this proposed approach. 

Deep Neural Networks (DNNs) have exhibited impressive performance in recent years \cite{8656578, 9011428}. Their success depends on high-quality labels, and a massive amount of data \cite{he2016deep}. However, obtaining high-quality data annotation is expensive and time-consuming because the labels of the dataset are all manually annotated. Therefore, some annotators choose to use the non-manufactured, semi-manufactured or online survey methods to improve the data annotation efficiency \cite{yan2014learning, yu2018learning, blum2003noise}.

In order to address the aforementioned noisy label issue, plenty of algorithms have been proposed for Learning with Noisy Labels (LNL) \cite{arpit2017closer, zhang2021understanding}. MentorNet \cite{jiang2018mentornet} has two networks, one is the student network, and the other is the pre-trained network to select the clean instances. However, the direction of the model update mainly depends on the selected clean instances from the teacher module, which is not robust enough. Inspired by \cite{jiang2018mentornet}, Co-teaching \cite{han2018co} is proposed for selecting clean data for updating the neural network. It trains the peer networks synchronously but with random initialization. However, the peer networks may converge rapidly due to the agreement approach. Co-teaching+ \cite{yu2019does} is proposed to further improve the performance and address the fast-fitting issue in Co-teaching with the disagreement algorithm. Few disagreed instances are utilized for parameter update even under the deeply corrupted annotation situation. In \cite{wei2020combating}, contrastive learning and joint loss are applied to address the noisy labels issue. However, applying only one approach to classify the noisy instance, especially with the high-level noise rate, is not robust enough due to the overfitting problem.

Hence, \textit{\textbf{Jo}}int \textit{\textbf{Co}}operative training with \textit{\textbf{T}}rinity Networks (JoCoT) is proposed to further enhance the LNL algorithm robustness. Specifically, it trains a trinity network that includes two teacher modules and one student module. The consensus outputs of the two teacher modules are fed into the student module to guide the clean instances mining. To verify the robustness of the proposed JoCoT, experiments are conducted on a widely used fall classification dataset, UP-Fall\cite{martinez2019up}, with different types of noises. The empirical results of JoCoT confirm that its robustness outperforms many other state-of-the-art noise tolerant approaches. The contributions of our proposed work are:

\begin{itemize}
\item A consensus-based method to mitigate the misclassification situation in the noisy label learning task.
\item A joint training framework to select the clean instances with high confidence.
\item A privacy mitigation in noisy labels learning with human skeleton limited body landmarks. 
\item To the best of our knowledge, the first evaluation of state-of-the-art methods on the skeleton data from the widely used UP-Fall dataset is also provided in this work.
\end{itemize}

\section{Related Work}
\subsection{Privacy Mitigating Fall Detection}
Due to the hazardous nature of fall events, the fall detection systems are needed for assisted living. The multi-view fall detection proposed in \cite{shu2021eight} could detect the fall action at a high altitude with a low time cost. However, the privacy protection is one of the main concerns of fall detection approaches. Using non-instrusive sensors to capture human motion information is one of the most widely used methods to protect the subject's privacy, e.g. human skeleton data obtained from surveillance cameras. In \cite{jeong2019human}, researchers proposed the skeleton feature extraction algorithms for fall detection. It exploits the relationship between different body parts of the subject and calculates the moving speed of the key body points between neighbourhood frames to detect fall events. Skeleton data were also applied in \cite{alaoui2019fall}, the skeleton data was first extracted from the original RGB data and then fed into the classifiers. The framework proposed in \cite{lin2020framework} applied LSTM, Gated Recurrent Unit (GRU) and Recurrent Neural Network (RNN) for learning the changes of the human body key points continuously. In \cite{chen2020fall}, three critical parameters were calculated for fall detection: the width-to-height ratio of subjects, the angle between the center-line and the ground of the subjects, and the speed of descent at the center of the hip joint. However, most of the works are focused on binary fall detection rather than multiple fall events classification. Aforementioned, a privacy-protected fall classification system based on the vision method is needed for a convenient healthcare system.

\subsection{Noise Tolerant Algorithms}
\subsubsection{Small loss}
In recent years, it has been proven that the DNN model will learn the simple and clean data first, and then the remaining data \cite{ren2018learning}. The clean data also have been proven to have a smaller loss value than the noisy data. Inspired by this theoretical foundation, small-loss is proposed for LNL, which selects the instances with small-loss as the clean data for training. MentorNet \cite{jiang2018mentornet} was proposed to first pre-train a teacher network and select the clean instances to ensure the correct updating direction for the student network. Co-teaching \cite{han2018co} selects the instances with small-loss as the clean data to update the parameters. 

\subsubsection{Data agreement and disagreement}

\begin{figure*}[h]
\centering
\includegraphics[width=14.4cm, height=6cm]{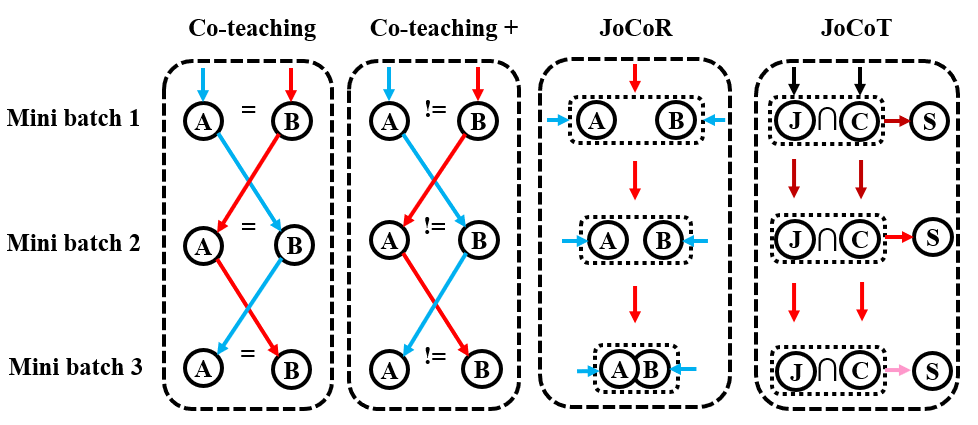}
\caption{The comparisons of the noisy label learning algorithms between Co-teaching\cite{han2018co}, Co-teaching+\cite{yu2019does}, JoCoR\cite{wei2020combating} and the proposed JoCoT. The selected errors are from the prediction errors, which are predicted by the peer networks A and B in Co-teaching, Co-teaching+ and JoCoR. For the JoCoT, since there exist both Co-teaching and JoCoR modules inside, therefore we assume the errors are from those two modules which are J and C. First panel: Co-teaching maintains two networks (A\&B). The parameters of the two networks are cross-updated with the agreement (=). Second panel: Co-teaching+ also maintains two networks (A\&B). Cross-updated is also applied in the peer networks but using disagreement (!=). Third panel: JoCoR maintains two networks (A\&B) and uses the joint loss function which contains both the contrastive loss and classification loss to make the predictions closer to each other. Fourth panel: JoCoT contains two teacher modules (J\&C) which denote the JoCoR and Co-teaching respectively, and apply the consensus ($\cap$) of the modules predictions to train the student module (S) to make the predictions of the corrupted data closer to the ground truth labels. The lighter colour indicates the data contain less noisy instances. (Best viewed in colored version)}
\end{figure*}
The theory of data agreement is to use the peer networks for selecting the instances that are supposed to be the clean data. As applied in Co-teaching \cite{han2018co}, two networks are trained in parallel, each network selects the small-loss instance as the clean instance with knowledge for information exchanging to the other network for training. Since different classifiers have different abilities for the decisions, which leads the individual bias for the performance. Data agreement could help the networks to find out the faults in the peer network. However, since the DNN model overfitted the noisy data with the increased training epochs \cite{han2018co}, the robustness of Co-teaching did not remain stable when the annotations were heavily corrupted.

The disagreement algorithm was applied to keep the peer networks diverge and increase the effective training stage length. Disagreement was first proposed in \cite{malach2017decoupling}, which calculated the loss value of the mini-batch instances, which have different predictions from the peer networks. It used the disagreed classified instances to guide the networks to train with the noisy data. In order to steer the networks to classify the corrupted data precisely and improve the performance, Co-teaching+ \cite{yu2019does} was proposed for combining the cross-update with disagreement. It could address the peer networks convergence issue and keep the network diverged.

\subsubsection{Other deep learning approaches}
\begin{table}[h]
\centering
\caption{Comparison of state-of-the-art algorithms with our proposed JoCoT method.}
\small\addtolength{\tabcolsep}{-1pt}
\begin{tabular}{c|c|c|c|c}
\hline
\hline
& Co-teaching & Co-teaching+ & JoCoR & \textit{JoCoT}\\
\hline
Small Loss & \checkmark& \checkmark & \checkmark & \checkmark \\
\hline
Cross Update & \checkmark& \checkmark & \xmark & \checkmark \\
\hline
Joint Training &\xmark &\xmark&\checkmark & \checkmark\\
\hline
Agreement &\xmark &\xmark &\checkmark &\checkmark\\
\hline
Disagreement &\xmark &\checkmark &\xmark &\xmark\\
\hline
Consensus &\xmark &\xmark &\xmark &\checkmark\\
\hline
\hline
\end{tabular}
\end{table}

Apart from the aforementioned approaches, several deep learning-based algorithms exist for training with the noisy label issue. Confident learning \cite{northcutt2021confident} proposed a noise pruning method to purify the corrupted dataset. The mathematical foundation is to estimate the dataset noise rate by counting the probabilistic thresholds and applying the confidence ranking in selecting the clean instances for training. PENCIL \cite{yi2019probabilistic} first initialized the distribution of the labels and applied supervised learning by the labels initialization distribution. With the training epochs increased, the label distribution was updated by back-propagation to estimate the noisy label distribution. 

The aforementioned disadvantages and limitations of current research motivated us to propose a robust deep learning system that can address noisy skeleton data issues for fall events classification.
We list the comparisons of state-of-the-art approaches with the proposed JoCoT in Table 1. It could be easily observed the proposed JoCoT exploits the strategies of the other approaches.

\section{Proposed Method}
In this section, detailed explanations of the proposed JoCoT are presented. We train a trinity network to tackle the noisy label issue. As shown in Figure 1, both teacher modules instruct the main network to mine the reliable and clean instances from the corrupted dataset for the training stage. Each mini-batch of the corrupted instances are simultaneously fed into the teacher modules and the prediction indexes for the noisy annotations are obtained. According to the predictions from the teacher modules, a consensus-based data selection strategy is applied for choosing reliable and clean data. After that, the selected data is fed into the student network for mining the valuable information for the parameter update. The consensus algorithm is updated in each training iteration of both teacher modules.

\begin{figure*}[]
  \centering
  \includegraphics[width=14.1cm, height=7.9cm]{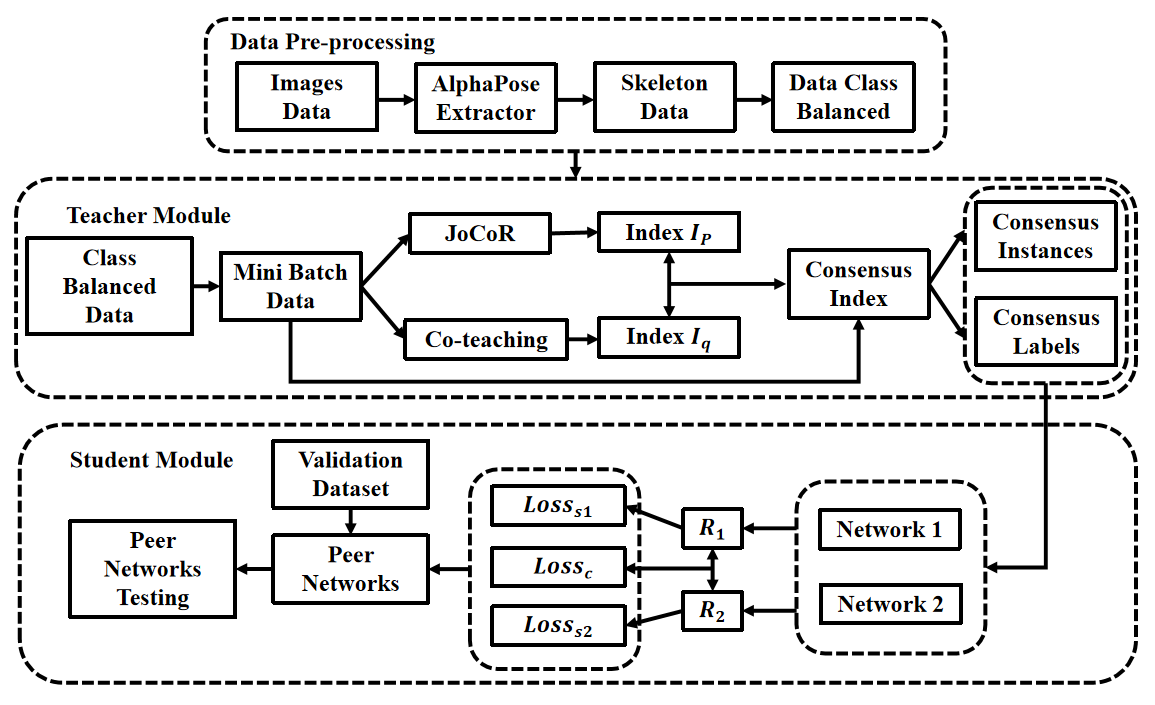}
\caption{The schematic of the proposed JoCoT. It is divided into three modules, data pre-processing, teacher module and student module. Firstly, the original RGB image data will be fed into the AlphaPose extractor \cite{fang2017rmpe} to obtain the human skeleton data, each sample contains 17 body landmarks. Since the original dataset is imbalanced, we re-scale the skeleton data to make sure each class of the activities contains the same number of samples. Secondly, the skeleton data will be fed into the teacher modules, which have JoCoR and Co-teaching for predicting the corrupted instances and apply the proposed consensus method to obtain the consensus `clean' instances and labels. Finally, those instances and labels will be fed into the student module for the network parameters updating. We apply the joint loss function between the predictions $R_{1}$ and $R_{2}$ from the peer networks to mine the noisy instances, which is precisely described in equation (2). The validation dataset is also supported for guiding the network updating towards the correct direction. }
\end{figure*}

\subsection{Preliminaries}
In order to verify the proposed algorithm JoCoT, we define the multi-class dataset as $D = \left \{ \textbf{\textit{x}}_{i}, y_{i} \right \}_{i=1}^{N}$, $N$ indicates the number of instances in the dataset. For the data training, $\textbf{\textit{x}}_{i}$ denotes the tensor quantity of $i$-th instance and the corrupted annotation $y_{i} \in \left \{1,2,...,M\right \}$, $M$ indicates the classes of human activities. Both of the teacher modules have the peer networks, which are denoted as $F(\textbf{\textit{x}}, \mathbf{\Theta_{1}})$, $F(\textit{\textbf{x}}, \mathbf{\Theta_{2}})$ and $G(\textit{\textbf{x}}, \mathbf{\Phi_{1}})$, $G(\textit{\textbf{x}}, \mathbf{\Phi_{2}})$, respectively. For the instance $\textbf{\textit{x}}$, $\forall \textbf{\textit{x}} \in D_{n}$, $D_{n}$ is the mini-batch dataset which is fetched from $D$.
Moreover, the prediction probabilities of the instances $\boldsymbol{x}_{i}$ from the two teacher modules are denoted as $\boldsymbol{p}_1=\left[p_1^1, p_1^2, \ldots, p_1^M\right], \boldsymbol{p}_2=\left[p_2^1, p_2^2, \ldots, p_2^M\right]$ and $\boldsymbol{q}_1=\left[q_1^1, q_1^2, \ldots, q_1^M\right]$, $ \boldsymbol{q}_2=\left[q_2^1, q_2^2, \ldots, q_2^M\right]$, which could also be considered as the ``softmax" layer outputs from the network parameters $\mathbf{\Theta_{1}}$, $\mathbf{\Theta_{2}}$, $\mathbf{\Phi_{1}}$, $\mathbf{\Phi_{2}}$, respectively.

For the proposed approach, both teacher modules could individually predict the annotations but train the network to update the parameters simultaneously with the peer paradigm. The cross-entropy loss $\mathcal{L}_{1}$ which applied in the first teacher module is as follows:
\begin{equation}
\begin{aligned}
     \mathcal{L}_{1}(\textbf{\textit{x}}_{i}, y_{i}) = \mathcal{L}_{s}(\textbf{\textit{x}}_{i}, y_{i})
     &= \mathcal{L}_{s_{1}}(\textbf{\textit{x}}_{i}, y_{i}) +\mathcal{L}_{s_{2}}(\textbf{\textit{x}}_{i}, y_{i})\\
     &=-2\boldsymbol{e}_{y_{i}} \log f(\textbf{\textit{x}}_{i})
\end{aligned}
\end{equation}
where $\mathcal{L}_{s_{1}}$ and $\mathcal{L}_{s{2}}$ indicate the sub-losses from the first teacher module. The classifier function $f$ maps the feature $\textbf{\textit{x}}_{i}$ to the label space and is defined as $f(\cdot): \mathcal{X} \rightarrow \mathbb{R}^{M}$, $\boldsymbol{e}_{y_{i}}$ is a one-hot vector which equals to $1$ if the predicted label is the same as $y_{i}$, otherwise equals to $0$. For the second teacher module, there exists the joint-training stage. The precise loss function $\mathcal{L}_{2}$ for $x_{i}$ is defined as followed:
\begin{equation}
    \mathcal{L}_{2}(\textbf{\textit{x}}_{i}, y_{i}) = (1-\lambda)\mathcal{L}_{1}\left (\textbf{\textit{x}}_{i},y_{i} \right )+\lambda\mathcal{L}_{c}(\textbf{\textit{x}}_{i})
\end{equation}
where $\mathcal{L}_{s}$ and $\mathcal{L}_{c}$ denote the supervised loss and contrastive loss functions in the joint loss function. Moreover, $\lambda$ is the weight parameter. The range of $\lambda$ is between 0.05 and 0.95, it is empirically selected for different noise rates. The explanation of the detailed supervised classification loss function is introduced in the following sub-section.

\subsection{Classification Loss}
Since the algorithm is proposed for addressing the multi-classification noisy label learning, we chose cross-entropy loss as the supervised classification loss function for all the peer networks from the teacher modules. It is widely used in multi-class classification tasks for calculating the loss value between the annotations and the predictions. Since the small-loss selection could help the network filter the noisy labels, we applied it with the cross-update theory \cite{han2018co} to reduce the annotation errors.
The complete supervised loss function $\mathcal{L}_{s}$ is defined as follows:
\begin{equation}
\begin{aligned}
    \mathcal{L}_{s}(\textbf{\textit{x}}_{i},y_{i}) &=\mathcal{L}_{s{1}}(\textbf{\textit{x}}_{i},y_{i})+\mathcal{L}_{s{2}}(\textbf{\textit{x}}_{i},y_{i}) \\
    &= -{\textstyle \sum_{i=1}^{N}}  {\textstyle \sum_{m=1}^{M}} y_{i}\log(q_{1}^{m}(\textbf{\textit{x}}_{i})) \\
    &-{\textstyle \sum_{i=1}^{N}}  {\textstyle \sum_{m=1}^{M}} y_{i}\log(q_{2}^{m}(\textbf{\textit{x}}_{i}))
\end{aligned}
\end{equation}
where $q_{1}^{m}(\textbf{\textit{x}}_{i})$ and $ q_{2}^{m}(\textbf{\textit{x}}_{i})$ denote the predicted probabilities of $i$-th instance from the peer networks for $m$-th label.

\subsection{Contrastive Loss}
According to the previous work \cite{han2018co}, the DNN model will tend to learn the simple and clean instances at the beginning of the training process. The peer networks will reach a consensus on most of the data but not on the corrupted data. In order to achieve better generalization ability, better performance, and guide the models to find more clean and reliable data. Co-regularization is constructed as the contrastive term in the loss function for the second teacher module, the same as in JoCoR. It could help the classifiers to maximize the agreement. Different from the traditional contrastive loss, the data augmentation is not required. One of the specific forms of Jensen-Shannon (JS), Kullback-Leibler (KL) divergence, is utilized to calculate the loss value between the peer networks' predictions. The symmetric KL divergence based contrastive loss function $\mathcal{L}_c$ is specifically defined as follows:
\begin{equation}
    \mathcal{L}_{c} = D_{KL}(\boldsymbol{q}_1\left |  \right | \boldsymbol{q}_2 )+D_{KL}(\boldsymbol{q}_2\left | \right |\boldsymbol{q}_1)
\end{equation}
The detailed definition of sub terms in $\mathcal{L}_{c}$ are as:
\begin{equation}
    D_{KL}(\boldsymbol{q}_1\left |  \right | \boldsymbol{q}_2)={\textstyle \sum_{i=1}^{N}}{\textstyle \sum_{m=1}^{M}}{q}_{1}^{m}(\textbf{\textit{x}}_{i})\log\frac{{q}_{1}^{m}(\textbf{\textit{x}}_{i})}{{q}_{2}^{m}(\textbf{\textit{x}}_{i})}
\end{equation}
\begin{equation}
    D_{KL}(\boldsymbol{q}_2\left |  \right |\boldsymbol{q}_1 )={\textstyle \sum_{i=1}^{N}}{\textstyle \sum_{m=1}^{M}}{q}_{2}^{m}(\textbf{\textit{x}}_{i})\log\frac{q_{2}^{m}(\textbf{\textit{x}}_{i})}{q_{1}^{m}(\textbf{\textit{x}}_{i})} 
\end{equation}
where $\boldsymbol{q}_1$ and $\boldsymbol{q}_2$ indicate the prediction probabilities of the peer networks. 
Hence, the negative memorization effects of noisy labels could be mitigated, and the classification performance could be further improved.

\subsection{Small Loss Selection}
The peer networks are more likely to reach a consensus on most instances but will have different predictions on the noisy data. According to the previous work focusing on the LNL algorithms \cite{han2018co, liu2015classification}, the instances with small loss are more likely to be clean. Therefore, the network model will be improved if the training data are with small-loss values. With the same decaying settings in JoCoR and Co-teaching, a parameter factor $R(T_{k})$ is defined to determine the proportion of the instances that should be selected related to the noise rate $\tau$. After each iteration, the small loss values will be sorted for selecting the mini-batch data. Due to the DNN model trends to learn the easy and clean instances at the beginning of the training process, but overfit to the noisy instances with the prolong of the training process, i.e., it has the best noisy instances filtering out ability at the beginning of the training process.
Therefore, in order to maximally keep the clean instances, $R(T_{k})$ will be gradually reduced to $1-\tau$ with the epochs increasing, i.e., $R(T_{k})$ is the largest at the beginning and then decrease. Hence, clean instances could be kept and noisy instances will be dropped before the network overfits to the noisy instances. The calculation of the remember rate $R(T_{k})$ is as follows.
\begin{equation}
    R(T_{k})=1-\min\left \{ \frac{T_{k}}{T}\tau,\tau  \right \} 
\end{equation}
where $T_{k}$ and $T$ indicate current epochs and total decay epoch numbers in the training stage. It could be observed that the $R(T_{k})$ will gradually decrease with the epochs increasing to keep the network trained with `clean' instances during the current training process. The detailed definitions of the instances selection algorithm in teacher modules are as follows:

\begin{equation}
    \bar{D}_{p}={argmin}_{D_{n}^{'}:\left | D_{n}^{'} \right |\ge R(t)\left | D_{n}\right | }\mathcal{L}_{1}(D^{'}_{n})
\end{equation}

\begin{equation}
    \bar{D}_{q}={argmin}_{D_{n}^{'}:\left | D_{n}^{'} \right |\ge R(t)\left | D_{n}\right | }\mathcal{L}_{2}(D^{'}_{n})
\end{equation}
where $\bar{D}_{p}$ and $\bar{D}_{q}$ indicate the small loss selection instances from the sorted mini-batch data $D^{'}_{n}$ in the teacher modules. In this case, the influences from the noisy data will be reduced.

\subsection{Consensus-Based Data Selection}
The framework of JoCoT contains three modules, including two teacher modules and one student module. Intuitively, the instances jointly selected by the two teacher modules are more reliable and clean-confident than those selected by a single teacher module. The teacher modules in the framework not only stabilize the selection process but also eliminate the noisy data, which helps to reduce the student module computational cost. To be precise, the two teacher modules will be trained in parallel, and the peer networks will select the clean instances in each mini-batch. We design a reasonable method to select reliable clean instances, which uses the consensus decision between the predictions of the teacher modules. Finally, the reliable clean instances will be fed to the student module to guide the training.

\begin{figure*}[h]
  \centering
  \includegraphics[width=15.1cm, height=5.1cm]{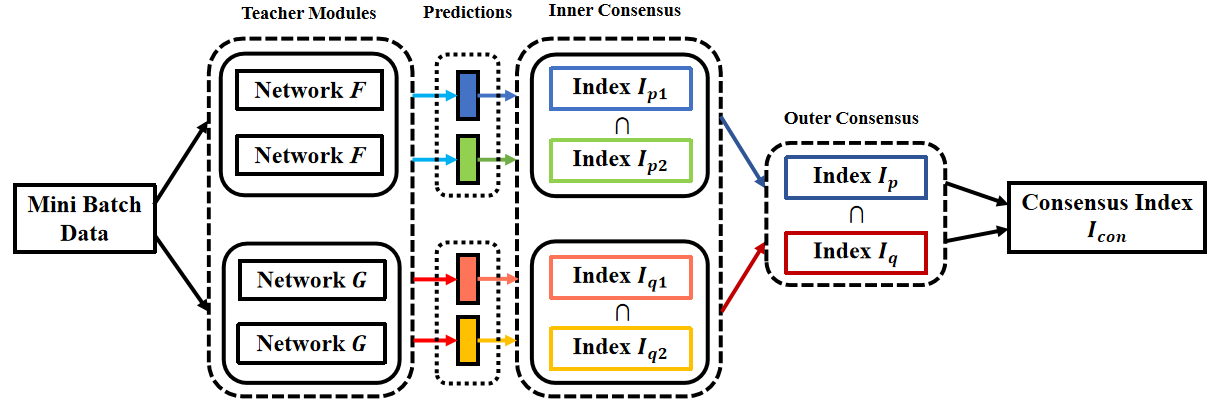}
\caption{The schematic of the proposed consensus method. Networks $F$ indicate the peer networks of JoCoR \cite{wei2020combating} and Networks $G$ indicate the peer networks of Co-teaching \cite{han2018co}. According to the four different colors predictions of the networks, the inner consensus will be obtained. Finally, the outer consensus data for both of the inner consensus indexes, $I_{p}$ for JoCoR and $I_{q}$ for Co-teaching will be obtained as $I_{con}$. This schematic will be repeated in each iteration until the end of the teacher modules training. (Best viewed in colored version)}
\end{figure*}
We demonstrate the framework of JoCoT in Fig. 2. The teacher modules which contain the peer networks are trained at the same time. After that, the consensus decision between the predictions from the teacher modules will guide the student module to classify the noisy data. A mini-batch of the input instances will be separately and simultaneously trained into the teacher modules. Each pair of networks in each teacher module will obtain a prediction index for the input data. Therefore, there will be four output indexes from the teacher modules. The consensus decision contains two steps which are called inner consensus and outer consensus. Inner consensus indicates that the consensus decision between the peer networks outputs from the same teacher module and outer consensus indicates the consensus decision between different teacher modules. The consensus-based data selection will be repeated in each data iteration for the teacher modules until the training stage is finished. After that, the consensus-based clean instances will be fed into the student module for human fall events classification. The detailed schematic of the proposed consensus method is shown in Fig. 3. This multi-step consensus data selection strategy could help the networks to utilize more robust and reliable clean instances to update the network parameters. The original data will be used as the testing set for both teacher modules and student module. The testing sets for the three modules are the same, and we use the student module outputs as the final results.

\begin{equation}
\begin{aligned}
I_{con} &= \left ( I_{\textbf{\textit{p}}_1}\cap I_{\textbf{\textit{p}}_2} \right ) \cap \left ( I_{\textbf{\textit{q}}_1}\cap I_{\textbf{\textit{q}}_2} \right )\\
&=  I_{\textbf{\textit{p}}}\cap I_{\textbf{\textit{q}}}
\end{aligned}
\end{equation}
where $I_{\textbf{\textit{p}}_1}, I_{\textbf{\textit{p}}_2}$ and $I_{\textbf{\textit{q}}_1}, I_{\textbf{\textit{q}}_2}$ indicate the output indexes from the teacher network modules $F$ and $G$ respectively. $I_{con}$ demonstrates each mini-batch final consensus-based selected data index. The calculation order of the equation will firstly be the inner brackets union (inner consensus) and then outside the brackets union (outside consensus). This procedure will continue until the end of the teacher training process. Afterwards, the selected instances and labels corresponding to the $I_{con}$ will be fed into the student module for training. The detailed process of JoCoT is shown in Algorithm 1.

\algnewcommand\algorithmicforeach{\textbf{for each}}
\algdef{S}[FOR]{ForEach}[1]{\algorithmicforeach\ #1\ \algorithmicdo}

\begin{algorithm}
\SetKwInOut{Input}{input}
\SetKwInOut{Output}{output}
\Input{Epoch $\mathcal{T}_{max}$, learning rate $\lambda$, iteration $\mathcal{N}_{max}$, sub networks $F, G$ with randomly initialization parameters $\mathbf{\Theta}_{1}, \mathbf{\Theta}_{2}$ and $\mathbf{\Phi}_{1}, \mathbf{\Phi}_{2}$;}
\BlankLine
\For{$t = 1,2,3,...,\mathcal{T}_{max}$}{
 \textbf{Shuffle} training set $D$\;
\For{$n=1,2,3,...,\mathcal{N}_{max}$}{
\textbf{Fetch} mini-batch $D_{n}$ from $D$;\\
$\textbf{\textit{p}}_{1}, \textbf{\textit{p}}_{2} = F(\textbf{\textit{x}}, \mathbf{\Theta_{1}}), F(\textbf{\textit{x}}, \mathbf{\Theta_{2}}) \ \forall \textbf{\textit{x}} \in D_{n}$;\\
\textbf{Calculate} loss value $\mathcal{L}_{1}$ using $\textbf{\textit{p}}_{1}$ and $\textbf{\textit{p}}_{2}$;\\
$\textbf{\textit{q}}_{1}, \textbf{\textit{q}}_{2} = G(\textbf{\textit{x}}, \mathbf{\Phi_{1}}), G(\textbf{\textit{x}}, \mathbf{\Phi_{2}}) \ \forall \textbf{\textit{x}} \in D_{n}$;\\
\textbf{Calculate} loss value $\mathcal{L}_{2}$ using $\textbf{\textit{q}}_{1}$ and $\textbf{\textit{q}}_{2}$;\\
\textbf{Obtain} the small-loss selections $\bar{D}_{p}$ and $\bar{D}_{q}$ by equation (8) and (9) from $D_{n}$;\\
\textbf{Update} the peer network parameters $\mathbf{\Theta}_{1}, \mathbf{\Theta}_{2}$ and $\mathbf{\Phi}_{1}, \mathbf{\Phi}_{2}$;\\ 
\textbf{Obtain} the inner consensused data index sets
$\boldsymbol{I}_p$, $\boldsymbol{I}_q$;
}

\textbf{Calculate} $\boldsymbol{I}_p \cap \boldsymbol{I}_q$ by equation (10);\\
\textbf{Obtain} the outer consensused data index set $\boldsymbol{I}_{con}$;\\

\textbf{Joint} both $\textbf{\textit{I}}_{con}$ and mini-batch data $D_{n}$ for the consensused-based clean instance set $\hat{D}_{n}$;\\
\textbf{Update} $R(t)$ by equation (7)
}
\Output{Consensused-based clean instance set $\hat{D}_{n}$}
\caption{JoCoT Teacher Modules Training}
\label{algo_disjdecomp}
\end{algorithm}

\section{Experimental Results}

\subsection{Dataset and Comparison Group}
\subsubsection{Dataset}
The custom dataset is based on the UP-Fall dataset. The whole dataset is collected in the same room environment. In the UP-Fall dataset, there are five fall events and six normal daily activities in both perspective cameras. The CAM1 is named as a sideway camera in the proposed work. To avoid confusion, we divided the dataset into 12 classes including one known activity. First, all the blank images without subjects were removed since the proposed approach aims at the individual-fall classification. This was implemented by applying the confidence score, and the highest confidence score was selected as the subjects in each frame.

Secondly, the human skeleton was selected as the data type in the training stage in order to mitigate the leakage of the subjects' privacy, which was caused by the background, dressing and facial information. It was obtained by using the pre-trained regional multi-person pose estimation network (RMPE) \cite{fang2017rmpe}, which is also well-known as AlphaPose. AlphaPose processes the RGB data from the UP-Fall dataset for extracting the human skeleton. The activity status of the subjects was presented by a feature vector which contains 17 key points. Each landmark contains three dimensions, which are the 2-D coordinates and confidence score.

Finally, since the proportion of positive and negative samples in the UP-Fall dataset is imbalanced, their ratio is approximately 3:97. Therefore, we rescaled the dataset by randomly selecting 1,200 samples from the original dataset in each activity. In total, there are 14,400 groups of skeleton data. We divided the dataset into three parts, training set, testing set, and validation set, and the ratio is 8:1:1. Validation set was applied to select the model with the best classification performance. Table 2 shows the details of the UP-Fall dataset, including the description and the numbers of each activity. The dataset has five fall events and seven normal human indoor activities. The number of the fall events is much smaller than the normal activities in the original training set. The dataset is comparatively clean and without a large amount of noisy data. In order to simulate the data corruption situation in the real world, noisy labels were added automatically to the dataset by using the noise-generating matrix which will be introduced in the following section. The imbalanced dataset problem significantly affects the actual noise rate far away from the noise rate settings and leads to a biased experimental performance. Rescaling the dataset could effectively address the noise generation problem.

\begin{table}[hbtp!]
\centering
\small\addtolength{\tabcolsep}{1.8pt}
\caption{Comparison between original and re-scaled training set and detailed description for UP-Fall dataset}
\begin{tabular}{c|c|c|c} \hline
Activity ID & Original & Re-scaled & Description\\ \hline
1 & 935 & 940& Falling using hands \\
2 & 922 & 958& Falling using knees \\
3 & 1,073 & 966& Falling backwards \\
4 & 902 & 942& Falling sideways \\
5 & 1,157 & 975& Falling to a chair \\
6 & 28,326 & 962 & Walking \\
7 & 39,762 & 964 & Standing \\
8 & 33,241 & 967 & Sitting \\
9 & 1,162& 955& Picking up objects \\
10 & 16,305 & 965 & Jumping \\
11 & 29,682 & 961 & Laying \\
12 & 995& 965 & Unknown \\ \hline
\end{tabular}
\end{table}

\subsubsection{Comparison Groups}
In this experiment, the JoCoT was compared with three different well known noisy label learning algorithms. The brief introductions of their theories are as followed:

\begin{enumerate}[label=(\roman*)]

\item  Co-teaching \cite{han2018co} simultaneously trained two peer deep neural networks and cross-updated the parameters of these networks by using the perceived clean instances, based on the small loss and agreement strategy.

\item  Co-teaching+ \cite{yu2019does} deployed the disagreement strategy and a cross-update algorithm based on small-loss selection was applied for training the peer deep neural networks.

\item  JoCoR \cite{wei2020combating} contained peer networks and updated the parameters by calculating the contrastive loss between the predictions of the peer networks.
\end{enumerate}

\subsection{Experimental Parameter Settings}
We applied a simple DNN network architecture both in two teacher modules and one student module. The experimental settings in both of student and teacher modules are the same. ReLU was used as the activation function. For the Adam optimizer used in all the experiments, similar to the settings in the baseline, we set the momentum as 0.9. Due to the skeleton data containing less information than the RGB  image data, the initial learning rates were set as 0.0001. The batch size was set to 128. The epochs were set as 300 for mining the clean instances in the consensus data selection stage. With the same learning rate decay point as in both teacher modules, the learning rate decayed gradually and linearly to zero from 80 to 300 epochs. The clean validation set guided the network training in the correct direction and prevented the over-fitting issue from noisy label instances. Different $\lambda$ were chosen to achieve the best performance under different noise rates and types. All the experiments were conducted on a workstation with 4 GeForce GTX 1080Ti GPUs and 16GB of RAM.


\subsection{Results and Discussions}
\subsubsection{Noise Types}
To simulate the real-world noisy dataset, we need to corrupt the dataset by using the noise-generating matrix $\mathcal{W}$. There are several noise types, e.g., pairflip and symmetric. Details of them will be introduced below, $\mathcal{P}$ denotes the noise rate and $M$ denotes the number of activities in the dataset:

\begin{enumerate}[label=(\roman*)]
    \item Pairflip flipping, a noise type which flips the ground truth to another specified activity among the entire dataset. The noisy matrix for pairflip is shown below:
    \\
    \begin{equation}
    \mathcal{W} = \begin{bmatrix}
    \centering
    1-\mathcal{P} & \mathcal{P} & 0 & \cdots & 0 \\
    0 & 1-\mathcal{P} & \mathcal{P} & \cdots & 0\\
    \vdots & \vdots & \ddots & \ddots &\vdots \\
    \vdots& \vdots & & 1-\mathcal{P} & \mathcal{P} \\
    \mathcal{P} & 0 & \cdots & 0 & 1-\mathcal{P} \\
    \end{bmatrix}
    \end{equation}
\\

    \item Symmetry flipping, which indicates that the noisy label is uniformly distributed over all labels except the true label, with an equal probability distribution. The detailed noise matrix explanation is shown below:
\\
    \begin{equation}
    \mathcal{W} = \begin{bmatrix}
    \centering
    1-\mathcal{P} & \frac{\mathcal{P}}{M-1} & \frac{\mathcal{P}}{M-1} & \cdots & \frac{\mathcal{P}}{M-1} \\
    \frac{\mathcal{P}}{M-1} & 1-\mathcal{P} & \frac{\mathcal{P}}{M-1} & \cdots & \frac{\mathcal{P}}{M-1}\\
    \vdots & \frac{\mathcal{P}}{M-1} & 1-\mathcal{P} &  &\vdots \\
    \vdots& \vdots &  & \ddots & \frac{\mathcal{P}}{M-1} \\
    \frac{\mathcal{P}}{M-1} & \frac{\mathcal{P}}{M-1} & \cdots & \frac{\mathcal{P}}{M-1} & 1-\mathcal{P} \\
    \end{bmatrix}
    \end{equation}
\\
\end{enumerate}

\subsubsection{Results on Pairflip}

To verify JoCoT performance at different levels of noise rate, we conduct the JoCoT with pairflip noise from 0.1 to 0.8. We introduce different noise levels in the experiments to simulate the corrupted dataset in the real-world environment.
\begin{table}[htbp!]
    \centering
    \begin{tabular}{c|c|c|c|c|c|c|c|c}
        \hline
        \hline
        Flipping-Rate & 0.1& 0.2& 0.3& 0.4& 0.5& 0.6& 0.7& 0.8  \\ \hline
        Co-teaching \cite{han2018co} & 87.16 & 85.16 & 76.51 & 75.10 & 47.88 & 32.84 & 13.45 & 8.50  \\ \hline 
        Co-teaching+ \cite{yu2019does}& 86.75 & 85.57 & 81.85 & 61.79 & 46.57 & 20.83 & 14.22 & 11.41 \\ \hline 
        JoCoR \cite{wei2020combating}& 88.18 & 86.12 & 83.52 & 78.71 & 44.32 & 25.63 & 12.03 & 11.04 \\ \hline 
        \textit{JoCoT} & \textbf{89.15} & \textbf{86.54} & \textbf{84.30} & \textbf{79.30} & \textbf{50.14} & \textbf{37.25} & \textbf{20.78} & \textbf{12.25} \\ \hline 
        \hline
        \end{tabular}
    \caption{Average test accuracy (\%) of Pairflip noise with different noise rates on UP-Fall.}
\end{table}

\begin{table}[htbp!]
    \centering
    \begin{tabular}{c|c|c|c}
            \hline
            \hline
            Flipping Level & Average & LR-Avg & HR-Avg  \\ \hline
            Co-teaching \cite{han2018co}& 53.33 & 74.36 & 18.26\\ \hline 
            Co-teaching+ \cite{yu2019does}& 51.12 & 72.51 & 15.49\\ \hline 
            JoCoR \cite{wei2020combating}& 53.69 & 76.17 & 16.23\\ \hline 
            \textit{JoCoT} & \textbf{57.46} & \textbf{77.89} & \textbf{23.43} \\ \hline 
            \hline
        \end{tabular}
    \caption{Average test accuracy (\%) of Pairflip noise with different noise rate levels on UP-Fall.}
\end{table}

As shown in Table 3, the test accuracy of Co-teaching and Co-teaching+ are not stable, Co-teaching+ shows better performance than Co-teaching with 30\% noisy data, the accuracy reaches 81.85\%. However, it drops significantly to 61.79\% when the noise rate increases to 40\%. Co-teaching achieves better performance than Co-teaching+ at 40\% noisy data, which is 75.10\% and without a drop but a steady decrease. JoCoT still achieves 79.30\% test accuracy with the same noise rate setting, which is 40\% of pairflip noisy labels. In a similar situation between JoCoR and Co-teaching+, it could be observed that the performance of JoCoR is better than Co-teaching+ at the low noise rate levels but trends to worse than Co-teaching+ when the rates are achieved 50\% and 70\%, which achieves 44.32\% and 12.03\%, respectively. It verifies that the robustness of the baseline approaches is not stable enough for responding to different noise rate settings. Although JoCoR achieves the best average performance in different levels of noise rate among the baseline approaches, it could not gain all the best performance with all the noise rate settings. Different from those baselines, the proposed JoCoT achieves the best performance of the others among all the noise rate levels. Besides these, we notice that all four algorithms have performance dropping when the noise rate reaches 50\%, and the accuracy decreases by at least 25\%. We assume the most important reason is that the density distribution of the pairflip noisy instances is imbalanced among all the activity categories. According to the definition of pairflip in equation (11), each of the activities only exists one specified noisy activity. Thus the density distribution of the pairflip noisy instances is imbalanced. Besides this, the noisy labels ratio of the corrupted dataset reaches a threshold (50\%), which indicates the noisy instances become more than half of the original clean dataset. The impact of the weight of the clean instances on the model is lower than the noisy instances, even though the DNN model first learns the clean and simple instances. Those factors can exacerbate the negative influence of the noisy samples, which leads to the incorrect parameter updating direction and the performance dropping.

According to Table 4, the average test accuracy of JoCoT is better than the other algorithms. It also presents the average testing accuracy in low noise rates (LR-Avg from 0.1 to 0.5 cases) and the average accuracy in high noise rates (HR-Avg from 0.6 to 0.8 cases). It could be observed that the performance improvement at HR-Avg is more significant than in the LR-Avg noise rate. The robust noisy label training method could mine valuable and clean instances even if the noise rate is relatively high. These confirm that the JoCoT has better noise tolerance and robustness to address the high-level noise rate issue for pairflip settings.

\subsubsection{Results on Symmetric}

Similarly, with the discussion in pairflip, Table 5 shows the performance of different algorithms with different symmetric noise rates. Among all the noise rate levels, it could be easily found that Co-teaching+ achieved the worst performance. The average performance for Co-teaching+ is 75.09\%. The LR-Avg and HR-Avg performances are much worse than the other algorithms, which are 82.43\% and 62.87\%, respectively. Regarding Co-teaching and JoCoR, JoCoR achieves better performance than Co-teaching from 10\% to 30\% noise rate. However, slightly lower than Co-teaching when the rates in 40\%, 50\% and 80\%. This same situation happened in symmetric noise, verifying that the baseline algorithms are not robust enough for the skeleton data-based noisy label learning on the UP-Fall dataset. Overall, the average JoCoR performance is 78.65\%, almost the same as the average performance of Co-teaching, which reaches 78.02\%. According to Table 5, JoCoT also achieves the best performance for symmetric noise among all the noise rate levels. When the noise rate achieves 80\%, JoCoT can also maintain the test accuracy at 57.35\% which indicates more than half of the data could be classified correctly. This verifies that the proposed JoCoT could effectively and robustly mine the clean instance under different noise rates. Moreover, according to Table 6, the average accuracy of JoCoT reaches 80.32\%, which is almost 5\% higher than Co-teaching+. The same in the pairflip section, the improvement of JoCoT obtained in HR-Avg is more than in the LR-Avg. This further verified the noise tolerant ability of JoCoT even under a high-level noise rate.
\begin{table}[htbp!]
    \centering
    \begin{tabular}{c|c|c|c|c|c|c|c|c}
        \hline
        \hline
        Flipping-Rate & 0.1& 0.2& 0.3& 0.4& 0.5& 0.6& 0.7& 0.8  \\ \hline
        Co-teaching \cite{han2018co}& 87.73 & 86.29 & 85.33 & 83.47 & 81.11 & 78.70 & 69.70 &  51.79 \\ \hline 
        Co-teaching+ \cite{yu2019does}& 85.77 & 84.65 & 84.31 & 80.16 & 77.24 & 73.56 & 68.35 & 46.69 \\ \hline 
        JoCoR \cite{wei2020combating}& 89.37 & 87.82 & 86.07 & 83.36 & 80.96 & 79.08 & 70.87 & 51.68 \\ \hline 
        \textit{JoCoT} & \textbf{89.44} & \textbf{88.10} & \textbf{86.32} & \textbf{84.49} & \textbf{82.55} & \textbf{80.63} & \textbf{73.71} & \textbf{57.35} \\ \hline 
        \hline
       \end{tabular}
    \caption{Average test accuracy (\%) of Symmetry noise with different noise rate on UP-Fall.}
\end{table}

\begin{table}[htbp!]
    \centering
    \begin{tabular}{c|c|c|c}
            \hline
            \hline
            Flipping Level & Average & LR-Avg & HR-Avg  \\ \hline
            Co-teaching \cite{han2018co}& 78.02 & 84.79 & 66.73\\ \hline
            Co-teaching+ \cite{yu2019does}& 75.09 & 82.43 & 62.87\\ \hline
            JoCoR \cite{wei2020combating}& 78.65 & 85.52 & 67.21\\ \hline
            \textit{JoCoT} & \textbf{80.32} & \textbf{86.18} & \textbf{70.56} \\ \hline 
            \hline
         \end{tabular}
    \caption{Average test accuracy (\%) of Symmetry noise with different noise rate levels on UP-Fall.}
\end{table}
In contrast to the pairflip performance change in Table 3, there is no performance dropping for symmetric in Table 5. The performance of all the algorithms decreased steadily with the increasing in the noise rate. This is due to the noisy instances being distributed uniformly in the dataset according to the definition of symmetric noise in equation (12). Therefore, different from the pairflip noise, the distribution of the noisy data is balanced. This could help to prevent the performance from dropping. JoCoT achieves the best performance among all the noise rate settings with symmetric noises, which justifies its clean instance mining and noise tolerant ability are better than the other approaches.

\subsection{Algorithm Robustness Analysis}
To further confirm the robustness of noisy data selection in different algorithms, we conduct experiments on the precision of the noisy data. Which indicates the number of truly noisy instances that the algorithms could find. The algorithm has better robustness and generalization ability if it could achieve higher precision of noisy labels, especially when the noise rate increases at a high level. 

\subsubsection{Pairflip analysis}

Table 7 indicates the precision performance for the noisy data with pairflip noise in the last training epoch, and Figure 4 shows the noisy label precision vs training epochs. Since we use part of the original dataset as our testing set, label precision experiments are conducted on the training set. According to Table 7, it could be significantly observed that the proposed JoCoT outperforms all the other baseline algorithms. The same with the test accuracy in Table 3, Co-teaching+ could find the least noisy instances on the average of all the noise rates. Comparing Co-teaching with Co-teaching+ in Table 7, we could observe that the precision of Co-teaching outperforms Co-teaching+ in most of the noise rates but not in 10\% and 50\%. This may be due to the disagreement applied in Co-teaching+ was to select the incorrect instances for mitigating the peer networks diverged, this may guide the parameter updating towards the erroneous direction. When the rate achieves 50\%, there exists a significantly dropping in all the approaches, as aforementioned, this is due to the noisy instances becoming over half of the dataset and misleading the training process.

\begin{figure*}[htbp!]
  \centering
  \includegraphics[width=14.9cm, height=4.4cm]{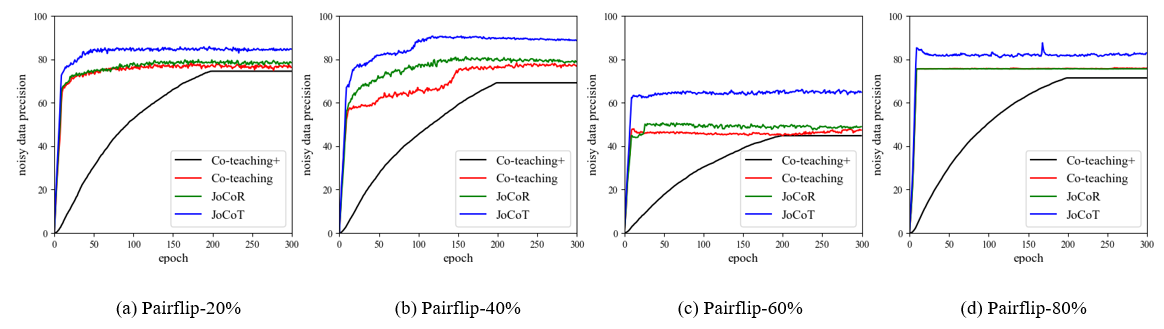}
\caption{Results on UP-Fall dataset with pairflip noise. Noisy label precision(\%) vs. epochs. The noisy label precision indicates the (true noisy data found by the algorithms) / (ground truth noisy data in total). (Best viewed in colored version)}
\end{figure*}

In Table 8, the average precision of Co-teaching+ reaches 65.19\% and 67.31\% for Co-teaching. Regarding JoCoR, the same situation happens. It could obtain 68.15\% on average, which is higher than Co-teaching, but the precision at 30\%, 70\% and 80\% noise rates are lower than Co-teaching.

Since JoCoR applied contrastivse learning loss between the peer networks predictions, we assume the reason for this is due to the skeleton data containing no such enough information for maintaining the JoCoR performance at some noise levels. This leads the peer networks from JoCoR to learn the incorrect information and mislead the direction of the parameter updating. No matter in the LR-Avg or HR-Avg, JoCoT always has significant improvement for finding the noisy instances and retains the clean instances for information mining. According to Table 8, JoCoT has a more significant improvement of precision at a high rate of noise than at a low rate, which reaches around 13\% improvement than Co-teaching+.
\begin{table}[htbp!]
    \centering
    \begin{tabular}{c|c|c|c|c|c|c|c|c}
        \hline
        \hline
        Flipping-Rate & 0.1& 0.2& 0.3& 0.4& 0.5& 0.6& 0.7& 0.8  \\ \hline
        Co-teaching \cite{han2018co}& 71.75 & 75.77 & 82.20 & 74.44 & 50.94 & 47.39 & 59.75 &  76.27 \\ \hline 
        Co-teaching+ \cite{yu2019does}& 73.03 & 74.51 & 78.33 & 69.17 & 52.70 & 44.81 & 57.55 & 71.41 \\ \hline 
        JoCoR \cite{wei2020combating}& 73.06 & 78.44 & 80.18 & 78.87 & 51.82 & 48.96 & 58.29 & 75.58 \\ \hline 
        \textit{JoCoT} & \textbf{81.26} & \textbf{85.35} & \textbf{84.78} & \textbf{87.01} & \textbf{60.99} & \textbf{61.46} & \textbf{74.40} & \textbf{85.64} \\ \hline 
        \hline
        \end{tabular}
    \caption{The precision of noisy data (\%) for Pairflip noise with different noise rates on UP-Fall.}
\end{table}

\begin{table}[htbp!]
    \centering
    \begin{tabular}{c|c|c|c}
        \hline
        \hline
        Flipping Level & Average & LR-Avg & HR-Avg  \\ \hline
        Co-teaching \cite{han2018co}& 67.31 & 71.02 & 61.14\\ \hline 
        Co-teaching+ \cite{yu2019does}& 65.19 & 69.55 & 57.92\\ \hline 
        JoCoR \cite{wei2020combating}& 68.15 & 72.48 & 60.94\\ \hline 
        \textit{JoCoT} & \textbf{77.61} & \textbf{79.97} & \textbf{70.62} \\ \hline 
        \hline
        \end{tabular}
    \caption{Average Noisy label precision (\%) of Pairflip noise with different noise rate levels on UP-Fall.}
\end{table}

Figure 4 shows the relation between the precision and epoch number with different noise rate settings. The final precision results of JoCoR and Co-teaching are all almost similar in the four figures. However, the changing in the curves are different. JoCoR has slightly increasing curves in both 20\% and 40\% of noisy data. However, in Fig 4(b), the precision of Co-teaching increases rapidly at the beginning of the training stage. The increasing speed becomes slow when the epochs arrive at 50. When it arrives at around 130, the precision increases rapidly again and reaches around 72\% at epoch 150 and reaches 74\%. The proposed JoCoT shows outstanding noisy data selecting ability with the best curves among the sub-figures in Fig 4.

Moreover, as shown in Tables 7 and 8, the precision of JoCoT achieves 77.61\% for the average precision, which achieves around 10\% improvement over JoCoR. Both the performances in LR-Avg and HR-Avg significantly outperformed all the other baseline algorithms, which confirms the robustness and noisy data mining ability of the proposed JoCoT in all the pairflip noise levels.

\subsubsection{Symmetric analysis}

Tables 9 and 10 show the precision of the algorithms with different noise rates of symmetric noise. Figure 5 demonstrates the algorithms' relationship between precision and epoch numbers under symmetric noises. Co-teaching+ shows the lowest precision among the noisy label learning methods under different noise rates. The average precision is 82.88\%. This may be due to the wrong clean instances selection of disagreement strategy applied in Co-teaching+ since the skeleton contains much less information than the information the RGB image contains. 

\begin{figure*}[htbp!]
  \centering
  \includegraphics[width=14.9cm, height=4.6cm]{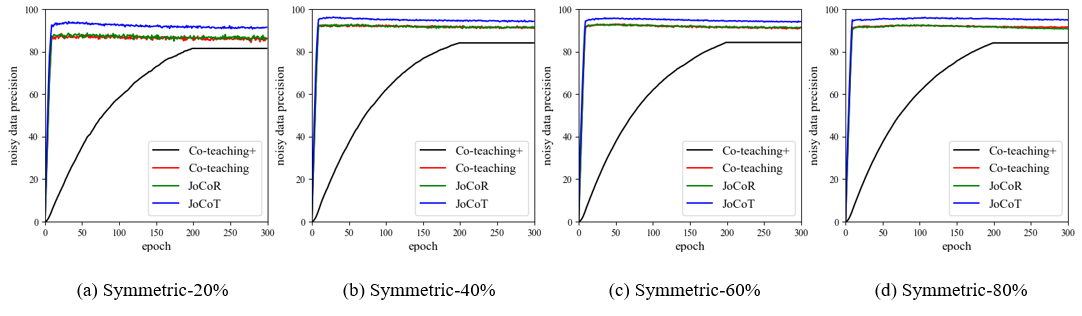}
\caption{Results on UP-Fall dataset with symmetric noise. Noisy label precision(\%) vs. epochs. The noisy label precision indicates the (true noisy data found by the algorithms) / (ground truth noisy data in total). (Best viewed in colored version)}
\end{figure*}

\begin{table}[htbp!]
    \centering
     \begin{tabular}{c|c|c|c|c|c|c|c|c}
        \hline
        \hline
        Flipping-Rate & 0.1& 0.2& 0.3& 0.4& 0.5& 0.6& 0.7& 0.8  \\ \hline
        Co-teaching \cite{han2018co}& 79.31 & 85.73 & 89.42 & 90.61 & 90.20 & 91.08 & 92.50 &  92.08 \\ \hline 
        Co-teaching+ \cite{yu2019does}& 77.71 & 81.51 & 82.88 & 84.10 & 83.44 & 84.32 & 84.96 & 84.09 \\ \hline 
        JoCoR \cite{wei2020combating}& 82.50 & 85.99 & 90.28 & 91.50 & 90.13 & 91.44 & 92.18 & 90.72 \\ \hline 
        \textit{JoCoT} & \textbf{88.72} & \textbf{91.22} & \textbf{93.46} & \textbf{94.24} & \textbf{93.24} & \textbf{94.64} & \textbf{95.30} & \textbf{95.97} \\ \hline 
        \hline
        \end{tabular}
    \caption{The precision of noisy data (\%) of Symmetry noise with different noise rates on UP-Fall.}
\end{table}

\begin{table}[htbp!]
    \centering
    \begin{tabular}{c|c|c|c}
        \hline
        \hline
        Flipping Level & Average & LR-Avg & HR-Avg  \\ \hline
        Co-teaching \cite{han2018co}& 88.87 & 87.05 & 91.89\\ \hline 
        Co-teaching+ \cite{yu2019does}& 82.88 & 81.93 & 84.46\\ \hline 
        JoCoR \cite{wei2020combating}& 89.34 & 88.08 & 91.44\\ \hline 
        \textit{JoCoT} & \textbf{93.35} & \textbf{92.18} & \textbf{95.30} \\ \hline 
        \hline
        \end{tabular}
    \caption{Average Noisy label precision (\%) of Symmetric noise with different noise rate levels on UP-Fall.}
\end{table}
Regarding JoCoR and Co-teaching, JoCoR could achieve higher precision than Co-teaching on average. However, as shown in Table 9, the JoCoR could not achieve better precision than Co-teaching among all the noise rates. For example, when the noise rate increases to 70\% and 80\%, Co-teaching reaches 92.50\% and 92.08\% respectively, but JoCoR achieves 92.18\% and 90.72\% respectively, which are lower than Co-teaching. As opposed to this, our JoCoT could consistently achieve the best performance. This confirms that our JoCoT could always find most of the noisy instances and verify its robustness. The average precision of JoCoT reaches 93.35\%. JoCoR has only 89.34\% for the averaged precision. Both LR-Avg and HR-Avg of JoCoT have nearly 4\% precision improvements than JoCoR. Each noise rate setting with JoCoT could achieve at least around 3\% improvement over JoCoR. In order to simultaneously analyze the performance in both Table 5 and Table 9, it could be observed that even if the noisy data precision of JoCoT achieves almost 95\% with 60\%-80\% symmetric noisy instances, which means over 95\% noisy data could be found, the accuracy of JoCoT still decreases from over 80\% to 57.35\%. This is because the number of noisy instances increases in percentage with the increasing of the noise rate. As aforementioned in the dataset setting section, there are 11,520 samples in the training set after the re-scaling operation. Even if the precision of JoCoT achieves around 96\% at 80\% noise rate. There still exist $11520\times(1-95.97\%)\times80\% \approx 371$ noisy instances in the dataset, which JoCoT does not select in the corrupted dataset. The noisy instances which are not found by JoCoT with a 10\% noise rate, which is approximately $11520\times(1-88.72\%)\times10\% \approx 130$ in the dataset. It is much smaller than the number of noisy instances at 80\% noise rate.
The more noisy instances that could not be found, the more significant the performance dropping occurs in the test accuracy. According to Figure 5. the precision of the JoCoT increases significantly at the beginning of the training stage and could maintain a high level with the increasing of the epochs. The results in Table 5, Table 9, Table 10 and Figure 5 further confirm that the noise tolerant and clean instances mining ability of JoCoT. 


The proposed JoCoT requires 13.76M bytes for the model parameters and 7.93 seconds for each epoch. Since the JoCoT contains both teacher and student modules, its computational cost is relatively large than the baseline approaches.

\section{Conclusions}

In this paper, we proposed a multi-class fall events detection approach called JoCoT to improve the performance of noisy label learning for fall event classification tasks. In JoCoT, a trinity network with two teacher modules and one student module was trained simultaneously. The consensus-based instances were applied from the teacher's peer networks for guiding the direction of the student network parameter updating. Comprehensive experiments were conducted on a widely used fall classification dataset called UP-Fall to verify the effectiveness of the JoCoT. Extensive experimental results demonstrate that JoCoT outperforms the state-of-the-art algorithms, which had 3.77\% and 1.67\% improvements compared with the state-of-the-art for pairflip and symmetric noise, respectively. For the high noise rates, JoCoT achieved 5.17\% and 3.35\% improvements compared with the state-of-the-art for pairflip and symmetric noise, respectively, which verified its robustness. In future work, the datasets which are recorded with a balanced cluster of demographics will be applied for better real-world generalization. Detailed generalization ability and domain adaptability are recommended for future work.
\bibliographystyle{ACM-Reference-Format}
\bibliography{LeiyuXie}
\end{document}